# Deep Fusion Prior for Plenoptic Super-Resolution All-in-Focus Imaging

Yuanjie Gu,[1,2] Yinghan Guan,[1,2,3] Zhibo Xiao,[1,2] Haoran Dai[2,4], Cheng Liu,[1,5] and Shouyu Wang[1,6,*]

[1]Computational Optics Laboratory, School of Sciences, Jiangnan University, Wuxi, Jiangsu 214122, China
[2]HorizonFlow Laboratory, China
[3]School of Electronic & Information Engineering, Chongqing Three Gorges University, Chongqing, 404020, China
[4]Department of Computer Science, Metropolitan College, Boston University, Boston, Massachusetts, 02215, USA
[5]Shanghai Institute of Optics and Fine Mechanics, Chinese Academy of Sciences, Shanghai 201800, China
[6]OptiX+ Laboratory, Wuxi, Jiangsu, China

**Abstract**. Plenoptic imaging offers not only 2-D projections but also adds light array directions, thus supporting single-shot all-in-focus imaging. While its poor spatial resolution becomes an obstacle to high-quality all-in-focus imaging performance. Although various super-resolution (SR) methods have been designed and combined with multi-focus image fusion (MFIF), high-quality multi-focus fused super-resolution images can be reconstructed for various applications, almost all of them deal with MFIF and SR separately. To our best knowledge, we first unify MFIF and SR problems as the multi-focus image super-resolution fusion (MFISRF) in the optical perspective and thus propose a novel dataset-free unsupervised framework named deep fusion prior (DFP) to address such MFISRF, particularly for plenoptic super-resolution all-in-focus imaging. Both numerical and practical experiments have proved that our proposed DFP approaches or even outperforms those state-of-the-art MFIF and SR method combinations. Therefore, we believe DFP can be potentially used in various computational photography applications. The DFP codes are open source and available at http://github.com/GuYuanjie/DeepFusionPrior.

*shouyu@jiangnan.edu.cn or shouyu29@icloud.com

## 1 Introduction

A majority of information acquisition, processing, and analysis is based on a visual perception system, which first records images, then distinguishes, recognizes, and extracts targets, and finally analyzes them to provide instructions for system decision and control. Among them, image recording is the premise for precision decisions. Common cameras only record 2-D projections, but lose depth information. In particular, when dealing with conditions with large depths of view, objects out of the depth of field of the imaging system become blurred and may induce errors in target distinguishing, recognition, and extraction. To break through the limit, all-in-focus imaging provides a solution. It first collects a multi-focus image stack and then fuses these multi-focus images into an all-in-focus one, where all objects are in focus [1]. According to the multi-focus image stack collection, there are mainly two tactics for all-in-focus imaging. One relies on the focus scanning of a common camera [2]. In other words, sliced images can be captured by moving the imaging system or scanning the focus along the optical axis. The advantage is that it can have high spatial resolution, but it sacrifices temporal resolution. Despite the use of many devices to accelerate focus scanning, such as electrically tunable liquid crystal lenses [3], micro-actuators [4], and optical diffusers [5], this method still cannot achieve single-shot all-in-focus imaging. Another tactic is based on numerically reconstructing a multi-focus image stack from a single-shot coded-aperture image [6]. Therefore, it completely solves the single-shot all-in-focus imaging problem. For example, plenoptic imaging is a representative technique. It offers not only 2-D projections but also adds light array directions. Therefore, it can reconstruct multi-focus images in color conditions and large fields of view and performs better than other coded-aperture-based techniques such as spectral focal sweep camera [7], color-filtered aperture [8], and lattice-focal lens [9]. Relying on many reported super-resolution (SR) techniques [10-15], and combined with multi-focus image fusion (MFIF) [16-20], plenoptic imaging is a promising tool to support super-resolution all-in-focus imaging in both high temporal and spatial resolution. Unfortunately, almost all of them deal with MFIF and SR separately. The separation inevitably induces complicated operations and long processing time.

Here, we propose deep fusion prior (DFP), which implements MFIF and blind SR with a unified, unsupervised, dataset-free, and robust model to deal with the MFISRF task, especially for plenoptic super-resolution all-in-focus imaging. We compare our DFP with the combinations of 3 MFIF methods and 3 SR methods. Experiments have proved that DFP approaches or even outperforms those state-of-the-art MFIF and SR method combinations. In addition, DFP has been successfully employed for plenoptic super-resolution all-in-focus imaging. In this manuscript, we first introduce the details of DFP in Section 2. Then, in Section 3, we compare DFP to other reported works in SR and MFIF. Next, we test the performance of DFP in plenoptic super-resolution all-in-focus imaging in Section 4. Finally, we conclude this work in Section 5.

## 2  Deep Fusion Prior

*2.1 Unified Optical Model for MFIF and SR*

Various MFIF and SR methods have been designed; however, almost all of them deal with MFIF and SR separately. However, we find that MFIF and SR share a unified optical model. The imaging process can be described as Eq. (1).

$$i(x,y) = h(x,y) * o(x,y) + n(x,y) \tag{1}$$

$h(x,y)$ is the 2-D point spreading function (PSF), $o(x,y)$ is the object, $n(x,y)$ is the additive noise, $*$ is the spatial convolution, and $i(x,y)$ is the image. For SR, its purpose is to use $i(x,y)$ to obtain an estimate $\hat{o}(x,y)$ of the real object $o(x,y)$.

While for MFIF, its model can be described in Eq. (2). $i_f(x,y)$ and $i_b(x,y)$ are the unfused images focusing on foreground and background. $m_f(x,y)$ and $m_b(x,y)$ are the decision maps determining the foreground and background regions. $E$ is the identity matrix and $u(x,y)$ is the fused image.

$$u(x,y) = m_f(x,y) \cdot i_f(x,y) + m_b(x,y) \cdot i_b(x,y) \tag{2}$$

$$m_f(x,y) + m_b(x,y) = E \tag{3}$$

Both $i_f(x,y)$ and $i_b(x,y)$ can be represented by Eqs. (4) and (5), in which $o_f(x,y)$ and $o_b(x,y)$ are the separated foreground and background objects from $o(x,y)$, and $h_f(x,y)$ and $h_d(x,y)$ are the 2-D in-focus and defocus PSFs.

$$i_f(x,y) = h_f(x,y) * o_f(x,y) + h_d(x,y) * o_b(x,y) + n_f(x,y) \qquad (4)$$

$$i_b(x,y) = h_d(x,y) * o_f(x,y) + h_f(x,y) * o_b(x,y) + n_b(x,y) \qquad (5)$$

Eq. (6) can be obtained by substituting Eqs. (4) and (5) into Eqs. (2) and (3), and it can be further generalized to Eq. (7). According to Eqs. (2) to (7), for MFIF, its purpose is to use $i_f(x,y)$ and $i_b(x,y)$ to obtain the $\hat{o}(x,y)$ for image fusion.

$$\begin{aligned} u(x,y) = \{ & \left[ m_f(x,y) \cdot h_f(x,y) + m_b(x,y) \cdot h_b(x,y) \right] * o_f(x,y) \\ & + \left[ m_f(x,y) \cdot h_d(x,y) + m_b(x,y) \cdot h_f(x,y) \right] * o_b(x,y) \} \\ & + m_f(x,y) \cdot n_f(x,y) + m_b(x,y) \cdot n_b(x,y) \end{aligned} \qquad (6)$$

$$u(x,y) = \hat{h}(x,y) * o(x,y) + \hat{n}(x,y) \qquad (7)$$

It reveals that MFIF and SR share a unified optical model. Therefore, MFIF and blind SR tasks can be combined as a multi-focus image super-resolution fusion (MFISRF) task. Based on this unified optical model, we design the DFP as a novel unified dataset-free unsupervised framework to address the MFISRF task, particularly for plenoptic super-resolution all-in-focus imaging.

*2.2 Deep Fusion Prior Framework*

Based on the unified MFISRF model mentioned above, the unsurprised DFP can obtain the high-quality fused super-resolution result using a single model, and it is trained with only two low-resolution input images without any external dataset. As revealed in Fig. 1, DFP consists of our designed SKIPnet end-to-end generated network, DoubleReblur tactic for focus measurement based on estimated PSF and Gaussian kernel convolution, decision embedding learned module for decision map optimization, and loss functions to guarantee high-quality MFISRF results robustly.

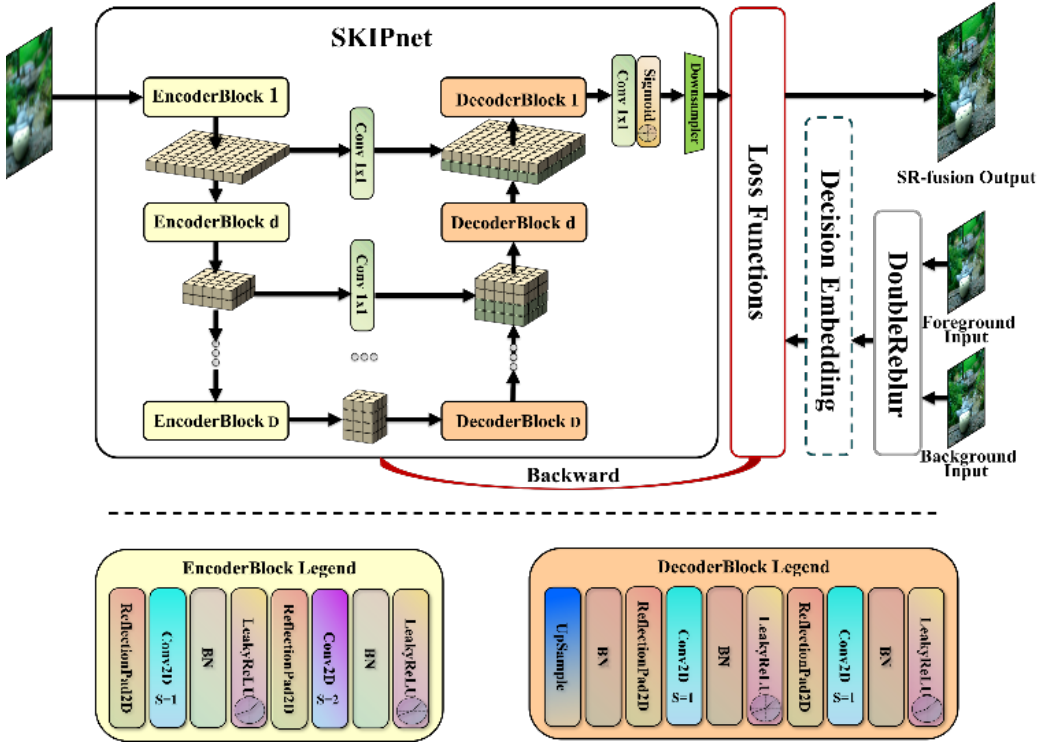

**Fig. 1.** DFP framework.

Based on the encoder-decoder framework, the SKIPnet is divided into the encoder and decoder parts, and they are symmetric about the central feature map $\phi_C$. The backbone of the encoder part is composed of $D$ encoder blocks that extract feature maps in $D$ scales. Each block consists of a reflection padding preparing layer, a 1-stride $n_d \times n_d$ convolution extracting layer, a batch normalization processing layer, a leaky-ReLU activating layer, a reflection padding preparing layer, a 2-stride $n_d \times n_d$ convolution downsampling layer, a batch normalization processing layer, and a leaky-ReLU activating layer successively. Identically, the backbone of the decoder part is composed of $D$ decoder blocks that extract and fuse feature maps in $D$ scales. Each block consists of a bilinear upsampling layer, a batch normalization processing layer, a reflection padding preparing layer, a 1-stride $n_d \times n_d$ convolution extracting layer, a batch normalization processing layer, a leaky-ReLU activating layer, a reflection padding preparing layer, a 1-stride $n_d \times n_d$ convolution extracting layer, a batch normalization processing layer, and a leaky-ReLU activating layer successively. For multi-scale feature map fusion, encoder feature maps $\phi \mathcal{D}_d^D(n)$ are concatenated to decoder feature maps $\phi \mathcal{U}_d^D(n)$. In the end, a 1×1 convolution aims at reducing the dimensionality. A sigmoid activation function is adopted to obtain the demanded output format. In the SKIPnet architecture, the downsampler with conventional approaches such as Bilinear, Bicubic, and Lanczos can obtain the same size output as inputs, and the scale of the SR depends on the scale of the downsampler. The depth D and the convolution kernel size nd are adjustable. For convenient parameter adjustment, $D = 5$ and $n_d = 5$ are used in DFP, and two 3×3 convolution layers can be used to replace the 5×5 convolution layer for higher efficiency.

Besides the SKIPnet, we also propose DoubleReblur for focus measurement and decision embedding for decision map optimization. Both their details are briefly mentioned in the Appendix. In addition, the designed loss functions as Eq. (8) are composed of content loss $L_{con}$, joint gradient loss $L_j$, and gradient limit loss $L_{grad}$. The content loss is to constrain the SKIPnet to learn the clear region of each image. The joint gradient loss is to enhance the SKIPnet to learn the high-frequency information and to reduce the dependency on decision maps. The gradient limit loss is to reduce the noise and oscillation effects. $\alpha$, $\beta$, and $\gamma$ in Eq. (8) are weighted parameters and set as 1, 0.5, and 0.1. More information on the loss functions can be referred to in the Appendix.

$$L = \alpha L_{con} + \beta L_{j\_grad} + \gamma L_{grad} \tag{8}$$

## 3 DFP verified by two-image super-resolution fusion

Before using DFP to deal with plenoptic super-resolution all-in-focus imaging, its performance in MFISRF was compared with reported techniques focusing on two-image super-resolution fusion. We qualitatively compared our proposed DFP with the combinations of learning-based MFIF (FusionDN [21], PMGI [22], U2Fusion [23]) and conventional SR (Bicubic)/unsupervised SR (DIP [24])/supervised SR (SRCNN [25]). These methods rely on the combinations of MFIF and SR to work with two models and learn from large datasets. While our proposed DFP only works with one model and learns without any dataset. Additionally, these methods, except DIP, are trained based on large datasets. However, DFP works without a dataset.

First, the DFP performance in MFISRF was tested. Fig. 2 reveals the visual results of MFISRF ×2 and ×4 on the MFI-WHU benchmark evaluation dataset [26]. In MFIF, PMGI improves the MFIF performance. It is compatible with multi-tasks such as infrared and visible image fusion, multi-exposure image fusion, and medical image fusion. But these tasks are often in low contrast cases, so the PMGI results are often color infidelity. While FusionDN and its upgraded U2Fusion compatible with multi-tasks have better MFIF performance than the above methods. However, they are still unable to fully resolve issues such as dark, blurred, and enhanced edges in MFIF. In SR, DIP improves image quality to some extent. But high-frequency details are lost in defocus and vista regions since only self-similarity and low-level statistical priors are used. Although SRCNN performs better than DIP, it relies on training with large datasets composed of fixed low and high-resolution image pairs.

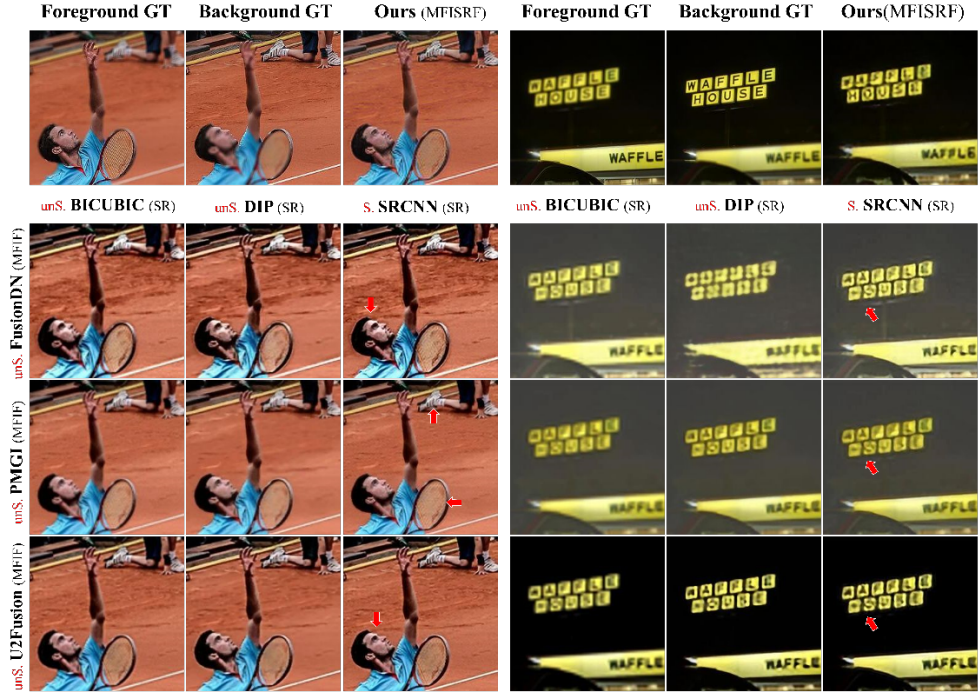

**Fig. 2**. Comparisons of our approach against multiple state-of-the-art supervised and unsupervised (S. & unS) learning-based MFIF (FusionDN, PMGI, and U2Fusion) and SR (Bicubic, DIP, and SRCNN) in ×2 and ×4 conditions on MFI-WHU 26 and MFI-WHU 08. In this figure, S. means supervised, unS. means unsupervised, and GT means ground truth.

Next, the DFP robustness was tested. Fig. 3 lists the results on MFI-WHU 29 with different DoubleReblur parameter sets. Fig. 3(A) shows the foreground and background ground truths. Fig. 3(B) provides the results obtained by DFP with the DoubleReblur parameter set as [3,5,5,0.05,1] or without weighted joint gradient loss. The result includes (1) the foreground decision map, (2) the MFISRF image, (3) the zoomed-in field of interest, and (4) the pseudo-color image depicting the difference between ground truth and MFISRF result in the Y channel. Identically, the images in Figs. 3(C), 3(D), and 3(E) are obtained by DFP but with different DoubleReblur parameter sets as [3,5,5,0,1], [3,5,5,0.01,1] and [3,5,5,0.05,1], respectively. These results prove that DFP has strong robustness since it can always provide high-quality MFIF results no matter whether the decision maps are good or bad.

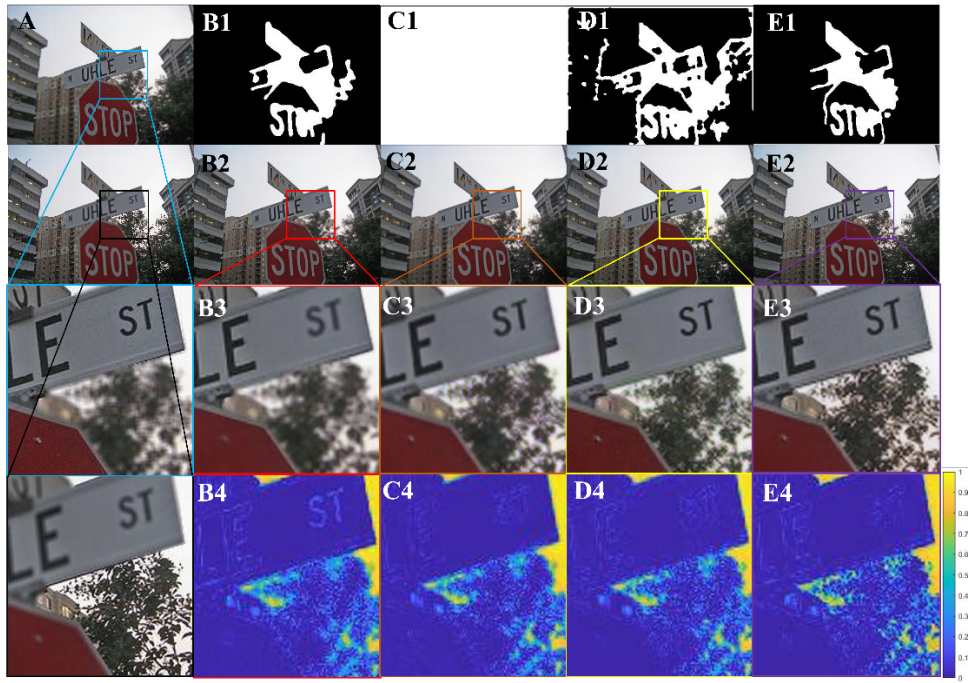

**Fig. 3.** Influence of decision maps on both MFIF and SR performances. (A) Foreground and background ground truths. (B) Results obtained by DFP with the DoubleReblur parameter set as [3,5,5,0.05,1] and without weighted joint gradient loss. (C), (D), and (E) Results obtained by DFP but with different DoubleReblur parameter sets as [3,5,5,0,1], [3,5,5,0.01,1] and [3,5,5,0.05,1], respectively. In (B)-(E), (1) is the foreground decision map, (2) is the MFISRF image, (3) is the zoomed-in field-of-interest, and (4) is the pseudo-color image depicting the difference between ground truth and MFISRF result in the Y channel.

Finally, quantitative comparisons are performed using the MFI-WHU benchmark evaluation dataset. The evaluation metrics include mean gradient (MG) [22], edge intensity (EI) [23], information entropy (IE) [26], mean gray value (MGA), and polar edge coherence (ECO) [27]. MG reflects the rate of contrast change of tiny details in the image. $MG_{(R)}$ describes the relative

MG between the ground truth (GT) and MFISRF. The most basic feature of an image is its edge existing between target and background, so it is one of the most important features to evaluate image fusion. EI is a quantitative coefficient to describe the edge information. $EI_{(R)}$ describes the relative EI between the GT and MFISRF. IE reflects the comprehensive characteristics of gray value at a pixel and its surrounding pixel gray distributions. $IE_{(R)}$ is the relative value between the GT and MFISRF. MGA is the average level of image gray, and it represents the overall brightness level of the image. $MGA_{(R)}$ describes the relative MGA between the GT and MFISRF. ECO reflects the edge similarity of the reconstructed and original images. ECO is edge coherence according to [27]. $ECO_{(R)}$ describes the relative ECO between the GT and MFISRF.

Tables 1 quantitatively compare the performances using different MFIF and SR combined methods and the proposed one according to the above-mentioned coefficients. Red marks the 1st best of the performance, and bold represents the 2nd and 3rd best. Our proposed DFP can achieve high-quality super-resolved multi-focus fused images. Even compared to the optimized combinations of unsupervised MFIF methods and supervised SR method, our proposed DFP could still obtain low $MG_{(R)}$, $EI_{(R)}$, $IE_{(R)}$, $MGA_{(R)}$, and $ECO_{(R)}$ values very close to or even lower than those obtained by the optimized MFIF and SR combined methods.

**Table 1. Comparison of our approach against multiple state-of-the-art learning-based MFIF and SR (BICUBIC, DIP and SRCNN)**[a]

| METHOD | $MG_{(R)}$ | $EI_{(R)}$ | $IE_{(R)}$ | $MGA_{(R)}$ | $ECO_{(R)}$ |
|---|---|---|---|---|---|
| **FusionDN+BICUBIC** | **0.9637** | **12.3182** | 0.3491 | 16.4175 | **0.0014** |
| **PMGI+BICUBIC** | 2.1311 | 18.4735 | 0.3899 | 18.1909 | 0.1422 |
| **U2Fusion+BICUBIC** | **0.9618** | 15.1490 | **0.2975** | **5.3168** | 0.0955 |

| Method | | | | | |
|---|---|---|---|---|---|
| FusionDN+DIP | 1.1171 | **14.3756** | 0.3529 | 16.7732 | 0.0158 |
| PMGI+DIP | 2.2334 | 19.5217 | 0.4057 | 18.5801 | 0.1329 |
| U2Fusion+DIP | 1.1756 | 17.2388 | **0.2929** | **5.2692** | 0.0940 |
| FusionDN+SRCNN | 1.8200 | 22.2081 | 0.3592 | 17.1748 | **0.0138** |
| PMGI+SRCNN | 1.7301 | 15.1158 | 0.3925 | 19.0188 | 0.1293 |
| U2Fusion+SRCNN | 2.1945 | 26.6641 | 0.3184 | 5.3172 | 0.0778 |
| DFP (ours) | **0.5840** | **6.3614** | **0.0222** | **0.5324** | **0.0056** |

[a] Red marks the 1st best of the performance, and bold represents the 2nd and 3rd best.

According to the above DFP verification on two-image super-resolution fusion, the proposed unsupervised dataset-free DFP approaches or even outperforms these state-of-art MFIF and SR method combinations. Additionally, DFP has strong robustness since it can always provide high-quality MFIF results less influenced by the decision maps. Therefore, these verifications prove that the proposed DFP works well in two-image super-resolution fusion.

## 4 DFP applied for plenoptic super-resolution all-in-focus imaging

After verification using two-image super-resolution fusion, our proposed DFP was applied for plenoptic super-resolution all-in-focus imaging. Fig. 4 reveals an example from the Lytro first generation dataset. Fig. 4(A) is the captured image, and Figs. 4(B)-4(E) are the reconstructed images focused at different planes via our previously proposed OpenRefocus [28]. Next, a plenoptic super-resolution all-in-focus image as shown in Fig. 4(F) was reconstructed from these multi-focus images via Bicubic. Additionally, another one revealed in Fig. 4(G) was also obtained via our proposed DFP. In comparison to these plenoptic super-resolution all-in-focus images, particularly those zoomed-in fields of interest, the proposed DFP not only has higher spatial resolution, but also provides a higher quality all-in-focus image, resulting in a better performance in the MFISRF task.

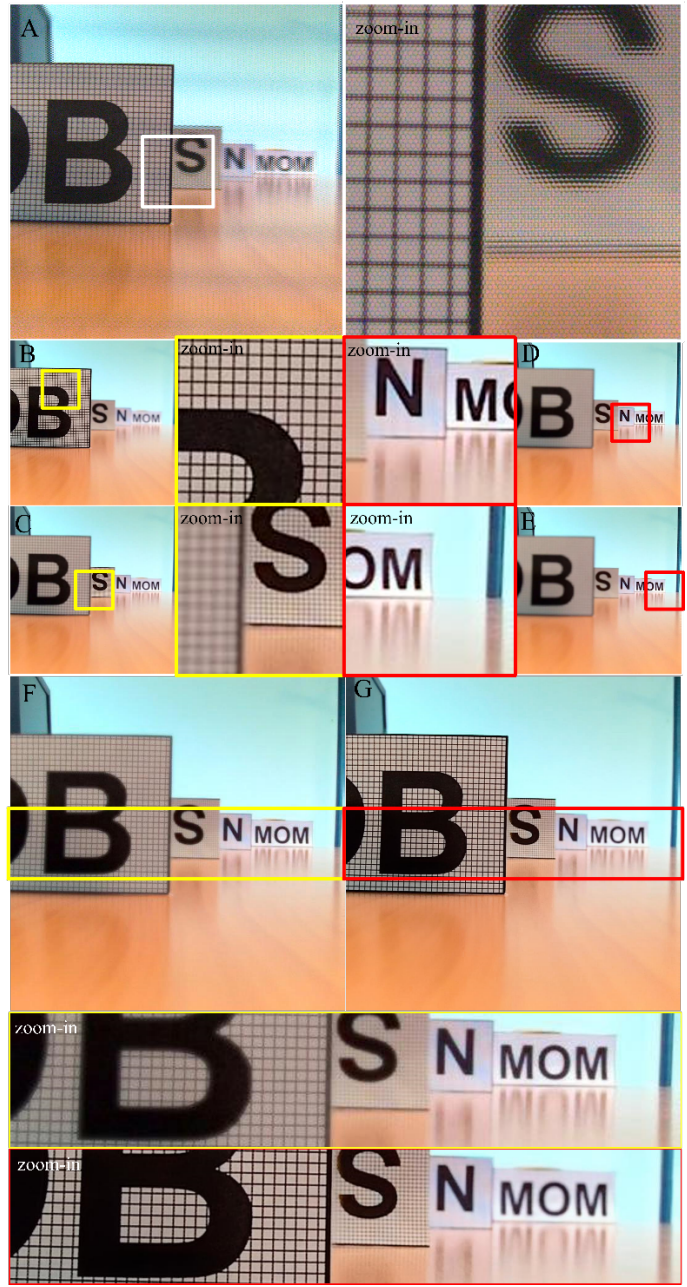

Fig. 4. Plenoptic super-resolution all-in-focus imaging example from the Lytro first generation dataset. (A) Captured image; (B)-(E) Multi-focus images; Super-resolution all-in-focus images reconstructed using (F) BICUBIC and (G) DFP.

Moreover, another practical experiment of capturing the complex natural light-field image using Lytro Illum was also provided in Fig. 5. Fig. 5(A) is our captured image of leaves at different depths of field. Figs. 5(B)-5(E) are the multi-focus images reconstructed using OpenRefocus. Figs. 5(F) and 5(G) are the plenoptic super-resolution all-in-focus images reconstructed using Bicubic and DFP, respectively. DFP, like the above example from the Lytro dataset, can obtain a higher spatial resolution and higher quality all-in-focus image in practice. Both these experiments verify that DFP can be successfully employed for plenoptic super-resolution all-in-focus imaging.

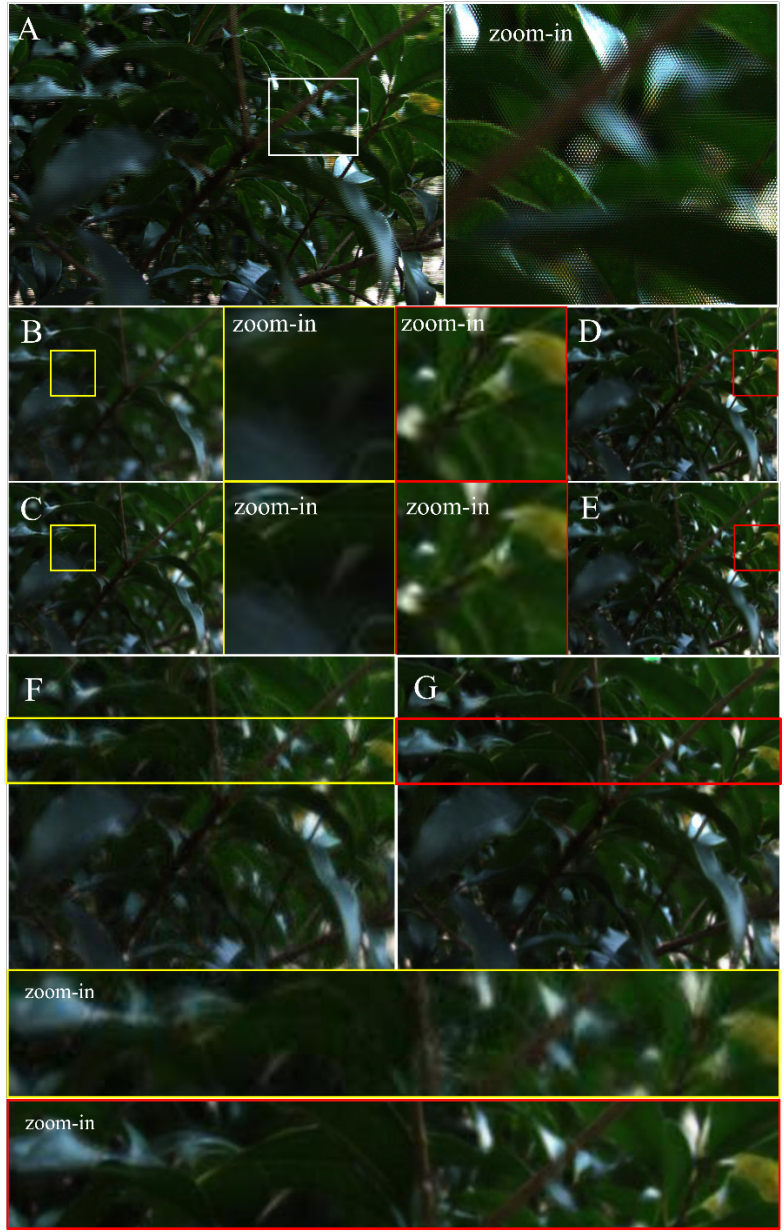

Fig. 5. Plenoptic super-resolution all-in-focus imaging example in practical application. (A) Captured image; (B)-(E) Multi-focus images; Super-resolution all-in-focus images reconstructed using (F) BICUBIC and (G) DFP.

## 5   Summary

In this study, we unify the MFIF and blind SR problems as the MFISRF task and propose a novel unified dataset-free unsupervised framework DFP for plenoptic super-resolution all-in-focus imaging. To the best of our knowledge, our proposed work is the first dataset-free unsupervised method to jointly implement the multi-focus fusion and super-resolution task for the first time. The DFP consists of our designed SKIPnet end-to-end generated network, DoubleReblur tactic for focus measurement based on estimated PSF and Gaussian kernel convolution, decision embedding learned module for decision map optimization, and loss functions to guarantee high-quality MFISRF results robustly. Compared to 3 MFIF and 3 SR method combinations, including both supervised and unsupervised ones, the proposed unsupervised dataset-free DFP approaches or even outperforms these state-of-the-art MFIF and SR method combinations. In addition, DFP has been successfully employed for plenoptic super-resolution all-in-focus imaging. Furthermore, DFP is a general framework, meaning that its networks and focus measurement tactics can be continuously updated to further improve the MFISRF performance. We believe DFP can be potentially used in various computational photography applications.

## 6   Appendix

*6.1 DoubleReblur*

Most MFIF methods are based on decision maps, which can be obtained by handcrafted focus measurement. Therefore, combining with computational imaging, morphological image processing, and graphics, we propose a DoubleReblur focus measurement model as shown in Fig. 6.

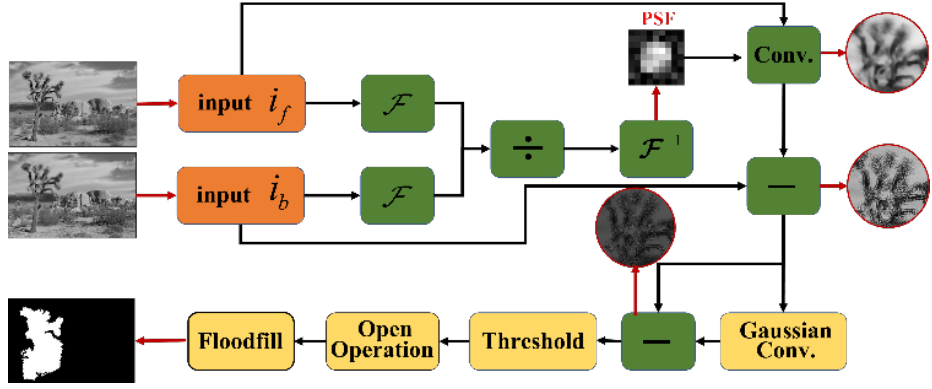

**Fig. 6.** DoubleReblur scheme.

According to the imaging model in Eq. (1), the relation between the foreground input $i_f(x, y)$ and the background input $i_b(x, y)$ can be represented as Eq. (9).

$$i_b(x,y) = i_f(x,y) * h_s(x,y) + n_s(x,y) \tag{9}$$

To estimate the spread kernel, Fourier transform is implemented on both sides of Eq. (9).

$$F\{i_b(x,y)\} = F\{i_f(x,y) * h_s(x,y) + n_s(x,y)\} \tag{10}$$

$$I_b(\xi,\eta) = I_f(\xi,\eta) \cdot H_s(\xi,\eta) + N_s(\xi,\eta) \tag{11}$$

In Eqs. (10) and (11), $F\{\cdot\}$ is the Fourier transform, $I_b(\xi,\eta)$, $I_f(\xi,\eta)$, $H_s(\xi,\eta)$, and $N_s(\xi,\eta)$ are the frequency spectra of $i_b(x,y)$, $i_f(x,y)$, $h_s(x,y)$, and $n_s(x,y)$ respectively.

$$\frac{I_b(\xi,\eta)}{I_f(\xi,\eta)} = \frac{I_f(\xi,\eta) \cdot H_s(\xi,\eta)}{I_f(\xi,\eta)} + \frac{N_s(\xi,\eta)}{I_f(\xi,\eta)} \tag{12}$$

To remove the noise term, a low-pass filter $T$ is performed on Eq. (12). In the end, the estimate of spread kernel $\hat{h}_s(x, y)$ can be obtained in Eq. (13).

$$\hat{h}_s(x,y) = F^{-1}\left\{T\left\{\frac{I_f(\xi,\eta)}{I_b(\xi,\eta)} - \frac{N_s(\xi,\eta)}{I_f(\xi,\eta)}\right\}\right\} \tag{13}$$

The first reblur image can be obtained by Eq. (14).

$$\tilde{i}_b(x,y) = i_b(x,y) * \hat{h}_s(x,y) \tag{14}$$

Then, the Gaussian reblur $\Gamma$ as the second reblur is used to enhance the sharpness difference as $s(x, y)$ obtained by Eq. (15).

$$s(x,y) = \left| \tilde{i}_b(x,y) - G\left\{\tilde{i}_b(x,y)\right\} \right| \tag{15}$$

In Eq. (16), $d(x, y)$ can be obtained by using threshold segmentation on sharpness difference image $s(x, y)$.

$$d(x,y) = \begin{cases} 1, & s(x,y) > t \\ 0, & s(x,y) \leq t \end{cases} \tag{16}$$

To eliminate gaps and holes, dilating and eroding are implemented as closed operation • in Eq. (17).

$$\hat{d}(x,y) = d(x,y) \bullet E \tag{17}$$

The decision map $m(x, y)$ can be finally obtained using the largest region floodfill algorithm demonstrated by Eq. (18).

$$m(x,y) = C\left\{\hat{d}(x,y)\right\} \tag{18}$$

There are 5 parameters in DoubleReblur as $[k_g, k_d, k_e, t, f]$: $k_g$ is the kernel size of Gaussian blur, $k_d$ is the kernel size of dilating, $k_e$ is the kernel size of eroding, $t$ is the threshold of segmentation, and $f$ is the bool flag control if the largest region filling is used. We transform the image to Y channel and use [5, 3, 3, 0.01, 1] for most images and slightly adjust parameters for different image details.

## 6.2 Decision Embedding

Since the decision maps are obtained from handcrafted focus measurement, they suffer from disadvantages such as edge fragmentation and false determination. To optimize the handcrafted decision maps, learning-based decision embedding is designed as Fig. 7. The input is obtained by the average of low-resolution inputs but can also be replaced by random noise. The binarized decision map can be generated by SKIPnet. Furthermore, as Eq. (19), we design an optimized loss function $L_{opt}$ to optimize the decision map by minimizing the difference between the focus measurement regions and the original low-resolution inputs.

$$L_{opt} = \frac{1}{H \cdot W} \sum_{i} \sum_{j} \left| \hat{m}_{i,j} - m_{i,j} \right| + \left| \hat{m}_{i,j} \cdot I_{fore_{i,j}} - I_{fore_{i,j}} \right| + \left| (1 - \hat{m}_{i,j}) \cdot I_{back_{i,j}} - I_{back_{i,j}} \right| \quad (19)$$

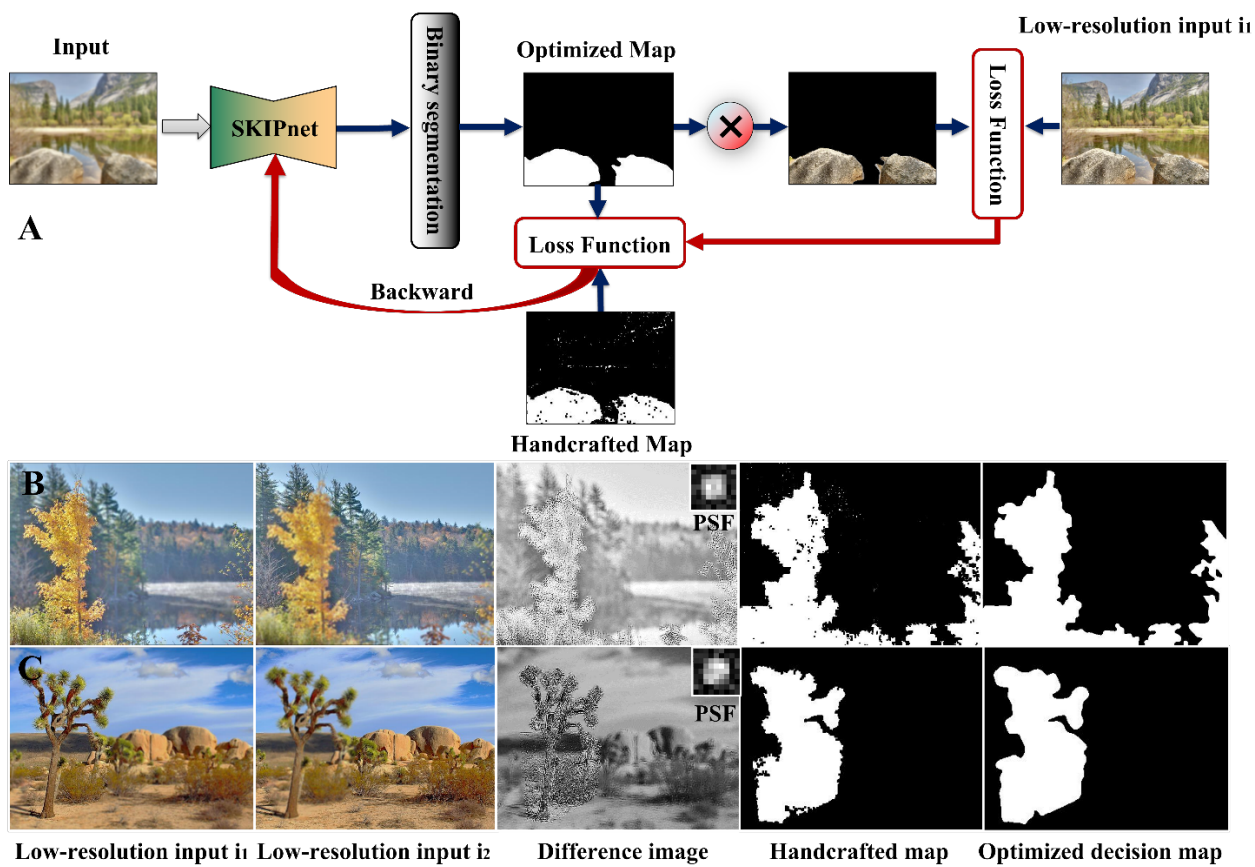

Fig. 7. (A) Decision embedding scheme and (B-C) examples.

In Eq. (19) $H$ and $W$ are the height and width of the image, $I_{fore}$ and $I_{back}$ are the foreground and background low-resolution inputs, $m$ is the handcrafted decision map obtained by DoubleReblur, $\hat{m}$ is the optimized decision map generated by SKIPnet. Algorithm 1 briefly demonstrates the decision embedding process. The decision embedding module is not indispensable, but the DFP performance will be better when considering it.

**Algorithm S1** Decision Embedding

1: Input: foreground input $I_{fore}$, background input $I_{back}$, input $n$ and handcrafted decision map $m$ which obtained by DoubleReblur;

Output: $\hat{m}$, $1-\hat{m}$.

2: **for** iterations K=500 **do**

3:     Forward: $\hat{m}$ = **SKIPnet**(input $n$);

4:     **for** x, y in $\hat{m}$ **do**

5:         **if** $\hat{m}(x,y) > 0.5$, $\hat{m}(x,y)=1$;

6:         **else** $\hat{m}(x,y)=0$;

7:     **end for**

8:

$$L_{opt} = \frac{1}{H \cdot W} \sum_{i} \sum_{j} |\hat{m}_{i,j} - m_{i,j}| + |\hat{m}_{i,j} \cdot I_{fore_{i,j}} - I_{fore_{i,j}}| + |(1-\hat{m}_{i,j}) \cdot I_{back_{i,j}} - I_{back_{i,j}}|$$

9:     Backward to update SKIPnet

       parameters Θ;

10:    **end for**

---

*6.3 Loss Functions*

$L_{con}$ in Eq. (20) is the content loss, where $\lambda_1$ and $\lambda_2$ are the weighted parameters both equal to 1. For generated MFISRF image $\hat{I}_{MFISRF}$, the focus regions in $I_{fore}$ and $I_{back}$ are obtained via decision maps. L$_1$ norms between the foreground focus region of $\hat{I}_{MFISRF}$ and $I_{fore}$, and between the background focus region of $\hat{I}_{MFISRF}$ and $I_{back}$ are used to compute their distances. Rather than *L$_1$* norm, *L$_2$* norm has stronger penalty for large errors and weaker penalty for small errors, and ignores the effect of the image content itself. It is worth noting the human visual system (HVS) is especially sensitive to brightness and color changes in untextured areas of the image.

$$L_{con} = \frac{1}{H \cdot W} \sum_i \sum_j \lambda_1 \hat{m} \cdot \left| \hat{I}_{MFISRF_{i,j}} - I_{fore_{i,j}} \right| + \lambda_2 (1-\hat{m}) \cdot \left| \hat{I}_{MFISRF_{i,j}} - I_{fore_{i,j}} \right| \quad (20)$$

$L_j$ in Eq. (21) is the joint gradient loss, where $\Delta_L$ is the gradient obtained via Laplacian. The difference between focus and defocus regions is almost the high-frequency information. Moreover, the Laplacian gradient map can well describe high-frequency information. For MFIF, the maximal Laplacian gradient map of unfused inputs almost equals to the fused image. For SR, the Laplacian gradient map can provide more high-frequency details. Therefore, *L*$_1$ norm between the Laplacian gradient map of generated MFISRF image $\hat{I}_{MFISRF}$ and the joint maximal Laplacian gradient map of low-resolution inputs $I_{fore}$ and $I_{back}$ is used to compute their distance.

$$L_{j\_grad} = \frac{1}{H \cdot W} \sum_i \sum_j \left| \Delta_L \hat{I}_{MFISRF_{i,j}} - \max(\Delta_L I_{fore_{i,j}}, \Delta_L I_{back_{i,j}}) \right| \quad (21)$$

$L_{grad}$ in Eq. (22) is the gradient limit loss. Unfortunately, DFP exploits the self-similarity property of input images themselves for reconstruction, thus inducing noise and oscillation. The gradient limit loss is used to reduce these noise and oscillation effects.

$$L_{grad} = \frac{1}{H \cdot W} \sum_{i} \sum_{j} (\nabla_x \hat{I}_{MFISRF_{i,j}} + \nabla_y \hat{I}_{MFISRF_{i,j}}) \qquad (22)$$

*Disclosures*

The authors declare no conflicts of interest.

*Acknowledgments*

The work was supported by National Natural Science Foundation of China (61705092) and Natural Science Foundation of Jiangsu Province of China (BK20170194).